\documentclass[conference]{IEEEtran}
\IEEEoverridecommandlockouts
\usepackage{cite}
\usepackage{amsmath,amssymb,amsfonts}
\usepackage{algorithm}
\usepackage{algorithmic}
\usepackage{graphicx}
\usepackage{textcomp}
\usepackage{xcolor}
\usepackage{hyperref}
\usepackage{subcaption}
\usepackage{tabularx, booktabs}
\newcolumntype{Y}{>{\centering\arraybackslash}X}
\def\BibTeX{{\rm B\kern-.05em{\sc i\kern-.025em b}\kern-.08em
    T\kern-.1667em\lower.7ex\hbox{E}\kern-.125emX}}

\begin{document}

\title{Capsule Neural Networks for Graph Classification using Explicit Tensorial Graph Representations\\
{\footnotesize }
\thanks{This work is part of an Innovate-UK funded research collaboration (GAMMA) between Braintree Ltd. (\url{http://www.braintree.com}) and University College London. Research grant no. 103971. This work has been submitted to the International Joint Conference on Neural Networks (IJCNN) 2019.}
}

\author{\IEEEauthorblockN{1\textsuperscript{st} Marcelo Daniel Gutierrez Mallea}
\IEEEauthorblockA{\textit{Research Lab} \\
\textit{Braintree Ltd.}\\
London, UK \\
m.gutierrez@braintree.com}
\and
\IEEEauthorblockN{2\textsuperscript{nd} Peter Meltzer}
\IEEEauthorblockA{\textit{Computer Science Dept.} \\
\textit{University College London}\\
London, UK \\
p.meltzer@cs.ucl.ac.uk}
\and
\IEEEauthorblockN{3\textsuperscript{rd} Peter J. Bentley}
\IEEEauthorblockA{\textit{Computer Science Dept.} \\
\textit{University College London}\\
London, UK  \\
p.bentley@cs.ucl.ac.uk}
}

\maketitle

\begin{abstract}
Graph classification is a significant problem in many scientific domains. It addresses tasks such as the classification of proteins and chemical compounds into categories according to their functions, or chemical and structural properties. In a supervised setting, this problem can be framed as learning the structure, features and relationships between features within a set of labelled graphs and being able to correctly predict the labels or categories of unseen graphs.

A significant difficulty in this task arises when attempting to apply established classification algorithms due to the requirement for fixed size matrix or tensor representations of the graphs which may vary greatly in their numbers of nodes and edges. Building on prior work combining explicit tensor representations with a standard image-based classifier, we propose a model to perform graph classification by extracting fixed size tensorial information from each graph in a given set, and using a Capsule Network to perform classification.

The graphs we consider here are undirected and with categorical features on the nodes. Using standard benchmarking chemical and protein datasets, we demonstrate that our graph Capsule Network classification model using an explicit tensorial representation of the graphs is competitive with current state of the art graph kernels and graph neural network models despite only limited hyper-parameter searching.
\end{abstract}


\section{Introduction}
Graph-structured data is prevalent in a broad set of domains including molecule representation, chemo- and bio-informatics, social network analysis, finance and many more. One reason for this being that graphical representations of data are able to model not only entities, but the connections and relationships between entities; thus offering a richer quality of information.

In order to develop successful machine learning models in these domains, we need techniques that can exploit this rich information inherent in the structure of a graph, as well as the entity feature information contained within a graph's nodes and edges \cite{morris2017glocalized}.

To define the scope of our work we adopt the distinction made in \cite{scarselli2009graph} between graph focused and node focused applications for machine learning on graphs. Examples of node focused applications include link prediction and node classification, where typical applications would include recommender systems and entity disambiguation. Whereas graph focused applications, which are the concern of this work, would include defining similarity, clustering or classifying graphs as instances themselves.

Many existing successful methods for graph classification are based on kernels \cite{kashima2002kernels} which are particularly well suited to the problem of comparing graphs of different dimensions; however, kernel methods with implicit representations suffer limited scalability \cite{kriege2014explicit}. Explicit kernels enable greater scalability, however present the challenge of mapping instances of various dimensions to a fixed size representation; and for graphs in particular, the challenge is increased by the permutation invariance requirement sought by a graph classifier. For example, given two isomorphic graphs in which the node ids in one are a permutation of the other, the graph classifier should recognise their equivalence.

We investigate the use of Capsule Networks \cite{sabour2017dynamic} to tackle the problem of supervised graph classification on undirected graphs of differing numbers of classes and discrete node features. Our hypothesis is that a Capsule Network would be better suited to identifying the similarities between graphs where permutations of the node ordering may cause other methods to fail, thus enabling greater classification performance. To the best of our knowledge, there has been no published literature combining explicit kernel graph representations with Capsule Networks. Our experiments demonstrate results that are competitive with other current state of the art methods on a set of seven widely used chemical and protein datasets.

To present our original contribution in applying Capsule Networks to graph classification through the use of explicit kernels, we first provide a review of related work in the graph classification domain. We then provide our methodology describing our contribution, with attention to both the tensor extraction and Capsule Network phases of our algorithm. This is followed by three experiments, first investigating the impact of different labelling procedures for the tensor extraction, second, bench-marking the performance of our complete model against current state of the art methods, and third, assessing the representation power of the Capsule Networks in the given tasks. After analysing our results we close with concluding remarks and discussion of potential further work.

The full source code for our work is available at  \href{https://github.com/BraintreeLtd/PatchyCapsules}{https://github.com/BraintreeLtd/PatchyCapsules}.

\section{Related Work}

\subsection{Graph Kernels}
Kernel methods have been successful in many machine learning applications \cite{hofmann2008kernel}, with many notable efforts in the graph classification setting \cite{gartner2003kernels,neuhaus2009kernel,vishwanathan2010graph}. A graph kernel is a positive semi-definite function defined on the space of graphs $\mathcal{G}$. This function corresponds to an inner product in some Hilbert space $\mathcal{H}$ to which the graphs are mapped: $\phi: \mathcal{G} \rightarrow \mathcal{H}$. The mapped space $\mathcal{H}$ may then be used with standard classification algorithms, or utilising the \emph{kernel trick} (with SVMs, for example) may be exploited implicitly. In this sense, kernel methods are well suited to deal with the high and variable dimensions of graph data, where explicit computation of such a feature space may not be possible.

Despite the high number of graph kernels in published literature, as distinguished in \cite{shervashidze2011weisfeiler}, they typically fall into just three distinct classes: Graph kernels based on walks and paths where random walks between two graphs are compared  \cite{borgwardt2005protein}, graph kernels based on a limited size subgraphs or graphlets, where the graph is represented by counts of subgraphs of different sizes \cite{shervashidze2009fast}, and graph kernels based on subtree patterns where a similarity matrix between two graphs is defined by the number of matching subtrees in each graph \cite{harchaoui2007image}.

Although graph kernels are well suited to produce good graph representations with respect to the difficulties in varying dimensions, the scalability of these graph kernels is limited; they scale poorly to large graphs. In the worst case, none of them scale better than $O(n^3)$ in the number of vertices \cite{shervashidze2011weisfeiler}.

One of the most notable graph kernels with respect to scalability in
the size of graphs is the kernel based on the Weisfeiler-Lehman (WL) algorithm for graph isomorphism \cite{weisfeiler1968reduction}. This kernel consists of repeatedly applying a hashing function to a node's neighbours' attributes and using the histogram of all the labels in order to represent the graph. It has attained state of the art results in terms of graph classification accuracy and in terms of execution time \cite{shervashidze2011weisfeiler}.

\subsection{Graph Neural Networks}
Several neural network models have been proposed for the problem of graph classification. 
The early works of \cite{gori2005new} and \cite{scarselli2009graph} present a Graph Neural Network (GNN) model based on information diffusion and relaxation mechanisms, and several instances of these recursive neural network models have been proposed since. Drawing upon the gains seen in the image classification domain with convolutional neural networks, \cite{bruna2013spectral} proposed a spectral graph convolutional neural network model that was later extended by \cite{defferrard2016convolutional} with fast localized convolutions. \cite{kipf2016semi} introduced a first order approximation of spectral convolutions on graphs with its graph convolutional network (GCN) model.

Generalising the processes involved in these graph convolutional networks, \cite{gilmer2017neural} defined a message passing neural network framework (MPNN) able to express many of the previous GNN models as specialised instances.
This framework defines two phases, a message passing phase where the hidden states of each node in the graph are updated according to the neighbours' messages, and a readout phase where a features vector is computed for the whole graph.

The GNN models described above have demonstrated state of the art results in label prediction \cite{kipf2016semi} and link prediction \cite{schlichtkrull2018modeling} problems. The GCN model works very well in label prediction tasks but it has problems with graph classification. Since the GCN provides node level outputs, to answer graph level questions requires some pooling process.
The main issue is that this model is equivariant with respect to the node order in a  graph \cite{verma2018graph}. This means that, given its non invariance to a permutation of the nodes, there are no guarantees that it would give the same results for two isomorphic graphs when the node ordering of the graphs is permuted, thus the pooling process may take different inputs for equivalent graphs.

One approach to solving this problem is proposed by \cite{verma2018graph} in adding a permutation invariant layer based on computing the covariance of the data.

So far, the design of the graph neural network models has been for the most part empirical and intuitive. \cite{xu2018powerful} showed theoretically that GNNs are at most as powerful as the WL isomorphism test in terms of distinguishing graph structures. They also showed that the first phase of the MPNN framework described in \cite{gilmer2017neural} can be divided into an aggregation and combination scheme followed by a readout function. The aggregation and the readout function need to be injective for a GNN model to be as powerful as the WL test.

It is interesting to note that even a powerful GNN model is bounded by the WL test in terms of discriminating graph structure; however the WL is limited to non continuous node features. A good GNN model satisfiying the injectivity conditions mentioned in \cite{gilmer2017neural} could potentially learn better representations in a graph classification problem.

\subsection{Experimenting with Both Worlds}

Graph kernels can be divided into explicit and implicit kernels. Implicit kernels compute a similarity measure between graphs and can benefit from the kernel trick. Explicit kernels compute a feature map for each graph directly; and, for large enough graphs, can be more efficient than implicit kernels \cite{kriege2014explicit}.

A model for learning explicit graph representations called Patchy-Sans was presented in \cite{niepert2016learning}. This algorithm extracts fixed size localized patches by applying a graph labeling given by the WL algorithm \cite{weisfeiler1968reduction} and the canonical labeling procedure from \cite{mckay2014practical}.
It then uses these patches to form a 3-dimensional tensor for each graph and uses a CNN to perform the classification.

Our experiments leverage a procedure similar to Patchy-Sans in order to convert a graph into a tensor representation. 
This representation is subsequently used in combination with a Capsule Network in order to perform graph classification. 
To date there is limited prior work in tackling graph classification problems with Capsule Networks, with  \cite{verma2018graph} being one example. The rationale for replacing the CNN with a Capsule Network is that there is a potential loss of information associated with the convolution operation. 

CNNs have been widely used in the machine vision community to address image classification problems \cite{krizhevsky2012imagenet}, however CNNs are only invariant to translation, and for this reason they cannot identify an object that has gone under a different transformation, such as a rotation.

Capsule networks work better with this problem by dividing the neurons into small groups in each network layer, where these groups are known as the capsules. The capsules correspond to concepts in different levels of abstraction during the process of parsing information. While this cross-layer association and the activation status of the capsules could represent semantic features in the case of image data, we expect that it would produce similar representations in the case of graphs \cite{lin2018learning}. 

To our knowledge, no previous work has combined an explicit kernel representation of a graph with a Capsule Network classifier. Our work fills this gap in order capture more accurately the structural information of a set of graphs. 

\section{Methodology}

In our experiments, we test the hypothesis that using a Capsule Network could help to address the permutation invariance problem in graph classification, and thus offer improved classification performance.

\subsection{Algorithm}
Following an introduction to the necessary notation, we present our contribution by considering the two distinct phases of our algorithm. The first phase generates a matrix representation for each graph in the dataset and the second applies a Capsule Network to these representations. We then provide the results and analysis of three experiments.

\subsection{Notation}
A graph $G_i \in \mathcal{G}$ is defined as a pair of vertices and edges $G =(V, E)$ of size $N = |V|$, where $V$ is the node set, $E$ the edge set, and $\mathcal{G}$ is the set of graphs of size $|\mathcal{G}|$.
For each graph $G_i$, it is possible to define the adjacency matrix between nodes as $A = [a_{i,j}]$ where $a_{i,j}$ is 1 if node $i$ is connected to node $j$ and 0 otherwise. Let $X \in R^{N \times d}$ be the node feature matrix, where $d$ is the dimension of the node features.

\subsection{Graph to Contextual Tensor}

In order to extract a tensor from a graph, the procedure described in Algorithm \ref{alg:g2m} is performed.
For each graph $G_i \in \mathcal{G}$, a node sequence order is defined following a given graph labelling procedure. The number of nodes that compose this node sequence is given by the width parameter $w$. Note that there is no requirement that adjacent nodes in the sequence are connected in the original graph.

For each node in this sequence, the closest neighbours (in terms of hop count) are gathered ($SubG_n$). The number of neighbours to gather is given by the height parameter $k$.
Then, a normalization of neighbours that follows a graph labelling procedure is used to order the labels by selecting nodes the give a sensible representation of the graph's structure. This enables discrimination between candidate nodes for selection in the case that there are too many nodes of equal hop count distance, as well as providing a consistent ordering within the subset of nodes selected. Finally, the categorical attributes of the nodes are encoded using one-hot encoding.


\begin{algorithm}[tb]
   \caption{GraphToTensor}
   \label{alg:g2m}
\begin{algorithmic}
\FOR{ \textbf{each} $G_i$ {\bfseries in} $\mathcal{G}$}
    \STATE {\bfseries Input:} Adjacency matrix $A_i$, node features $X_i$, width $w$, receptive field size $k$.
    \STATE NodeList $\gets$ NodeSequenceOrdering($A_i$, $X_i$,$w$)
    \STATE $M_i \gets$ Array[]
    \FOR{$n$ {\bfseries in} NodeList}
        \STATE $SubG_{n}$ $\gets$ NeighbourGathering($A_i$, $X_i$,$n$)
        \STATE $SubG_{n}$ $\gets$ Normalization($SubG_{n}$)
        \STATE $M_i[n] \gets SubG_{n} $
   \ENDFOR
   \STATE $T_i \gets $ Encoding($M_i$)
\ENDFOR
\end{algorithmic}
\end{algorithm}

This procedure generates a receptive field for each selected node in the graph.
Two different labelling procedures for the selection and ordering of nodes are investigated: a canonical labelling procedure, and a ranking based on betweenness centrality.
The purpose of using the canonical labeling procedure is to order the nodes in the graph so that they correspond to the isomorph class of a given graph, whereas the betweenness centrality procedure finds the most connected nodes in each graph. The benefit in using either of these procedures being to select nodes consistently across different graphs, thus providing the same representations for isomorphic graphs, and similar representations for similar graphs.

The resulting shape of the tensor representation $T_{i}$ is given by $w \times k \times d $  where $w$ is the width, $k$ the receptive field and $d$ is the number of dimensions of the one-hot encoded node features.

\subsection{Graph Capsule Network}
Once we have the tensorial representation of the graph $X_{tr}$, we apply a Capsule Network architecture with reconstruction as regularization method. 

\begin{figure}[ht]
\vskip 0.2in
\begin{center}
\centerline{\includegraphics[width=\columnwidth]{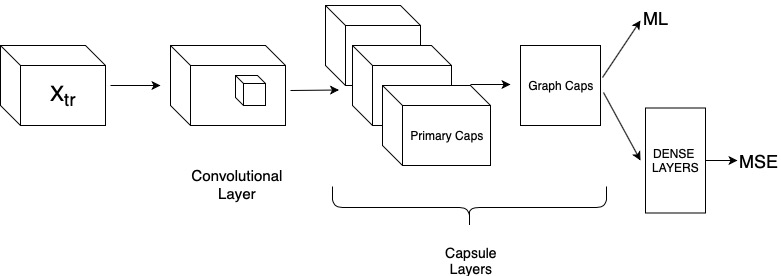}}
\caption{Graph Capsule Network Architecture}
\label{gcnArc}
\end{center}
\vskip -0.2in
\end{figure}

The architecture of the Capsule Network is depicted in \autoref{gcnArc}. The first layer is a CNN and the second (primary caps) and third (graphcaps) layers are capsule layers.
The main difference between a Capsule Network layer and a standard neural network layer is the existence of a routing-by-agreement procedure. This procedure is described in detail in \cite{sabour2017dynamic} but a brief description is provided below.

In \autoref{gcnArc}, the primary caps represent the first layer of the capsules network and there is no routing between these capsules and the first convolutional layer. The graph caps represent the second layer of the Capsule Network.
Between the primary caps and the graph caps a routing procedure is in place. This iterative procedure works as follows, each lower level capsule sends its input to the higher level capsule, if this capsule agrees with the lower one's input then this information will be backpropagated during training and strengthen the link between these capsules.

Finally, there is a decoder layer after the graph caps layer that acts as a regularizer. The total loss is given by 
\begin{equation}
\label{total_loss}
    L =  MSE + ML
\end{equation}
MSE is the loss that comes from reconstructing the graph using the decoder layer. It is the mean square error between the reconstructed graph and the original graph.

ML is the margin loss and it is defined as the categorical cross-entropy loss in the case of two classes or, in the multi-class classification case, it is defined as follows
\begin{equation}
\label{ml_loss}
    \begin{split}
    ML = \quad & T_k\cdot \max(0,m^{+} - |v_k|)^2 \\
         & + \lambda (1 - T_k)\cdot \max(0,|v_k|-m^{-})^2 
    \end{split}
\end{equation}
where $T_k = 1$ if and only if a graph class $k$ is present, and $m^+ = 0.9$ and $m^{-} = 0.1$. The $\lambda$ down-weighting of the loss for absent graph classes stops the initial learning from shrinking the lengths of the activity vectors of all the graph capsules.

\begin{table*}[t]
\caption{Graph Statistics}
\centering
\label{graph_stats}
\begin{tabularx}{\linewidth}{|l|Y|Y|Y|Y|Y|Y|Y|}
\hline
\multicolumn{1}{|c|}{\textbf{Dataset}} & \multicolumn{1}{c|}{\textbf{MUTAG}} & \multicolumn{1}{c|}{\textbf{PTC}} & \multicolumn{1}{c|}{\textbf{PROTEINS}} & \multicolumn{1}{c|}{\textbf{NCI1}} & \multicolumn{1}{c|}{\textbf{NCI109}} & \multicolumn{1}{c|}{\textbf{D \& D}} & \multicolumn{1}{c|}{\textbf{ENZYMES}} \\ \hline
\textbf{No. Graphs ($|\mathcal{G}|$)} & 188 & 344 & 1113 & 4110 & 4127 & 1178 & 600 \\ \hline
\textbf{Max. Graph Size} & 28 & 109 & 620 & 111 & 111 & 5748 & 126 \\ \hline
\textbf{Avg. Graph Size} & 18 & 25.56 & 39.06 & 29.8 & 29.6 & 284.32 & 32.6 \\ \hline
\textbf{Number of classes} & 2 & 2 & 2 & 2 & 2 & 2 & 6 \\ \hline
\textbf{Number of node labels (n)} & 7 & 18 & 3 & 37 & 38 & 82 &  \\ \hline
\textbf{Class ratio (Percentage of  + labels)} & 66.49\% & 39.51\% & 59.57\% & 50.05\% & 50.38\% & 58.66\% & 16.67\% \\ \hline
\end{tabularx}
\end{table*}

\begin{table*}[ht]
\caption{Time for training the models (seconds)}
\centering
\label{time_tab}
\setlength\tabcolsep{5pt}
\begin{tabularx}{\linewidth}{|X|r|r|r|r|r|r|r|}
\hline
\multicolumn{1}{|c|}{\textbf{Algorithm\textbackslash{}Dataset}} & \multicolumn{1}{c|}{\textbf{MUTAG}} & \multicolumn{1}{c|}{\textbf{PTC}} & \multicolumn{1}{c|}{\textbf{PROTEINS**}} & \multicolumn{1}{c|}{\textbf{NCI1*}} & \multicolumn{1}{c|}{\textbf{NCI109*}} & \multicolumn{1}{c|}{\textbf{D \& D*}} & \multicolumn{1}{c|}{\textbf{ENZYMES**}} \\ \hline
\textit{Nauty + Capsules} & 133.64 $\pm$ 4.59 & 184.9 $\pm$ 13.31 & 1255.6 $\pm$ 8.95 & 5191.23 $\pm$ 24.56 & 5221.51 $\pm$ 48.21 & 4034.25 $\pm$ 0.51 & 366.24 $\pm$ 2.18 \\ \hline
\textit{Nauty + CNN} & 13.03 $\pm$ 0.87 & 76.57 $\pm$ 3.83 & 70.2 $\pm$ 0.24 & 640.14 $\pm$ 15.73 & 1632.35 $\pm$ 24.85 & 1033.22 $\pm$ 2.15 & 32.71 $\pm$ 0.06 \\ \hline
\textit{BC + Capsules} & 133.64 $\pm$ 4.59 & 138.55 $\pm$ 4.23 & 1039.98 $\pm$ 3.25 & 5065.57 $\pm$ 33.32 & 5045.7 $\pm$ 39.49 & 3671.17 $\pm$ 0.66 & 366.24 $\pm$ 2.18 \\ \hline
\textit{BC + CNN} & 13.03 $\pm$ 0.87 & 53.37 $\pm$ 1.63 & 72.51 $\pm$ 2.3 & 603.29 $\pm$ 12.88 & 1598.87 $\pm$ 45.32 & 997.28 $\pm$ 1.94 & 32.71 $\pm$ 0.06 \\ \hline
\end{tabularx}
\end{table*}
 
\subsection{Representational Power} 
We use the t-SNE algorithm \cite{maaten2008visualizing} to visualize the high dimensional representations learned by the Capsule Networks in both the CNN and capsule layers, as well as directly on the explicit tensor representations we use as input to the networks.

This algorithm consists of approximating the distribution of the probability distribution of the distances between points in the high dimensional space to the probability distribution in the low dimensional space. The metric used to measure the similarity between both probability distributions is the Kullback-Leibler (KL) divergence \cite{hershey2007approximating}.

\subsection{Implementation Details}
Our implementation uses TensorFlow \cite{abadi2016tensorflow}, with the Adam optimizer \cite{kingma2014adam} and an exponentially decaying learning rate of $1e-6$ to minimize the sum of the margin losses in \autoref{total_loss}.

We perform a grid search with the Mutag dataset (see \autoref{graph_stats}), in order to find the optimal hyper-parameters using 10 fold cross-validation with a $90\%-10\%$ train-test split. We search on the following space:
\begin{itemize}
    \item Number of epochs = [100, 150, 200]
    \item Learning rate = [0.0005, 0.001, 0.005]
    \item Learning rate decay = [0.25, 0.4, 0.75, 1.5]
\end{itemize}
Note, we do not perform a hyper-parameter search on each dataset because the computational cost is too high. Instead we use the same set of hyper-parameters found for the Mutag dataset in each of our experiments.

\section{Experiments}

To investigate our hypothesis that the Capsule Networks are better suited to the permutation invariance problem of graph classification we conduct three experiments to test the two separate phases of the algorithm, and to measure the performance of our graph Capsule Network classifier for comparison against current state of the art methods.

The first stage of our algorithm generates explicit graph tensor representations which are processed downstream by models typically applied to images (CNNs + Capsule Networks) where the order of pixels in a given sample is of obvious significance. Since our model is designed to operate on graphs, where the order of nodes in a sample may be given in any order (i.e. for isomorphic graphs with permuted node ids), to ensure a fair test of our model we must guarantee that the samples we test on are in no way pre-ordered by any systematic method, whether deliberate, or by chance through manual curation of the datasets. To provide this guarantee, in all experiments we first randomly permute all node ids.

\subsection{Datasets}
 
The graph Capsule Networks are tested against the MUTAG, PTC, NCI1, NCI109, PROTEINS and D$\&$D. \autoref{graph_stats} summarises the important features of the datasets that are analyzed and a description of each dataset can be found below.
 
\subsubsection*{MUTAG}

This is the dataset of mutable molecules, it contains 188 chemical compounds, and it can be divided into two classes according to whether they are mutagenic or not, where 125 of them are positive and 63 are negative \cite{yu2017graph}.

\subsubsection*{NCI}

This collection of graph datasets is commonly used as the benchmark for graph classification. Each NCI dataset belongs to a bioassay task for anticancer activity prediction, where each chemical compound is represented as a graph, with atoms representing nodes and bonds as edges. A chemical compound is positive if it is active against the corresponding cancer, or negative otherwise \cite{pan2015cogboost}.

\subsubsection*{PTC}

This graph dataset includes a number of carcinogenicity tasks for toxicology prediction of chemical compounds. The dataset contains 417 compounds from four types of test animals: MM (male mouse), FM (female mouse), MR (male rat), and FR (female rat). Each compound is with one label selected from {CE, SE, P, E, EE, IS, NE, N}, which stands for Clear Evidence of Carcinogenic Activity (CE), Some Evidence of Carcinogenic Activity (SE), Positive (P), Equivocal (E), Equivocal Evidence of Carcinogenic Activity (EE), Inadequate Study of Carcinogenic Activity (IS), No Evidence of Carcinogenic Activity (NE), and Negative (N) \cite{pan2017task}. We performed the experiments in each of the datasets (MM, FM, MR and FR) and averaged the results.

\subsubsection*{PROTEINS AND ENZYMES}

These are sets of proteins from the BRENDA database \cite{schomburg2004brenda} and the dataset of Dobson and Doig \cite{dobson2003distinguishing}, respectively. Proteins are represented by graphs where nodes represent secondary structure elements (SSEs), which are connected whenever they are neighbors either in the amino-acid sequence or in 3D space. Each node has a discrete type attribute (helix, sheet or turn) and an attribute vector containing physical and chemical measurements including length of the SSE in Angstrøm ($\mathring{A}$), distance between the $C_{\alpha}$ atom of its first and last residue in A, its hydrophobicity, van der Waals volume, polarity and polarizability. ENZYMES comes with the task of classifying the enzymes to one out of 6 EC top-level classes, whereas PROTEINS comes with the task of classifying into enzymes and non-enzyme \cite{feragen2013scalable}.

\subsection{Experiment 1: Ablation Study and Comparison of Labelling Procedures}
To provide an ablation study and make the performance difference of a Capsule Network over a CNN in the latter phase of our algorithm explicit, we compare both classifier models here on seven common graph classification benchmarking datasets. We also compare two different labelling methods: Canonical Labelling using NAUTY \cite{mckay2014practical} and Betweenness Centrality \cite{brandes2001faster} to inform the node selection process in which the contextual tensors of each sample graph are generated.

\subsection{Experiment 2: Comparison Against Current State of the Art Methods}
We compare the results of our approach with current state of the art graph kernels and the graph neural networks methods for graph classification on the same graph classification benchmarking datasets.

\subsection{Experiment 3: Assessment of the Representational Power of the Capsule Network}
 
We compare the representational power of the different representations that are provided by the tensor extraction phase of our algorithm. The CNN has three layers, the input layer, the inner layer and the output layer. The Capsule Network has 5 layers, an input layer, a  convolutional layer, and primary capsule layer, a graph capsule layer and a decoder layer. We visualize the inner layer of the CNN and the primary capsule layer of the Capsule Network because, after the training procedure, these are the layers that contains a manifold (non-linear) representation of the graph.

For ease of analysis, here we focus on two sets of graphs on either ends of the graph size spectrum; one with a small number of nodes per graph (Mutag), and one with a large number of nodes per graph and (Proteins).


 
\subsection{Experimental Setup}
Experiments 1 and 2 were performed using two different hardware setups according to the sizes of the datasets.
For the Mutag, PTC, Proteins and Enzymes datasets we used a computer with 16GB memory size, 3.1 GHz Intel Core i7 CPU, and 8 cores.

The NCI1, the NCI109 and the D$\&$D datasets have a larger number of graphs and a larger number of nodes in each graph, this make them computationally more expensive. For these reason we used a a p2xlarge Amazon EC2 instance with 1 GPU with 12 GB of memory, 4 vCPUs and 64 GB of memory. The latter setup was also used to measure the execution time results shown in \autoref{time_tab}.


\section{Results}

\begin{table*}[t]
\caption{Ablation study - CNN vs Capsule Network classification accuracies, and comparison of labelling procedures}
\label{cnn_vs_caps}
\centering
\begin{tabularx}{\linewidth}{|l|Y|Y|Y|Y|Y|Y|Y|}
\hline
\multicolumn{1}{|c|}{\textbf{Algorithm\textbackslash{}Dataset}} & \multicolumn{1}{c|}{\textbf{MUTAG}} & \multicolumn{1}{c|}{\textbf{PTC}} & \multicolumn{1}{c|}{\textbf{PROTEINS}} & \multicolumn{1}{c|}{\textbf{NCI1}} & \multicolumn{1}{c|}{\textbf{NCI109}} & \multicolumn{1}{c|}{\textbf{D \& D}} & \multicolumn{1}{c|}{\textbf{ENZYMES}} \\ \hline
\textit{Nauty + Capsules} & 75.7 $\pm$ 9.47 & 63.9 $\pm$ 6.36 & 72.0 $\pm$ 2.61 & 59.4 $\pm$ 2.16 & 58.0 $\pm$ 2.76 & \textbf{77.9 $\pm$ 2.49} & 26.1 $\pm$ 5.15 \\ \hline
\textit{Nauty + CNN} & 85.2 $\pm$ 5.66 & 53.8 $\pm$ 6.47 & 70.4 $\pm$ 2.20 & 56.4 $\pm$ 2.09 & 58.0 $\pm$ 2.76 & 75.3 $\pm$ 4.44 & 22.3 $\pm$ 4.02 \\ \hline
\textit{BC + Capsules} & \textbf{88.9 $\pm$ 5.49} & \textbf{69.0 $\pm$ 4.98} & \textbf{74.1 $\pm$ 3.24} & \textbf{65.9 $\pm$ 1.07} & \textbf{58.04 $\pm$ 2.78} & 74.86 $\pm$ 3.27 & \textbf{27.0 $\pm$ 8.45} \\ \hline
\textit{BC + CNN} & 84.2 $\pm$ 5.26 & 57.6 $\pm$ 2.01 & 68.9 $\pm$ 3.38 & 57.6 $\pm$ 2.01 & 56.9 $\pm$ 2.03 & 72.3 $\pm$ 3.86 & 20.0 $\pm$ 5.57 \\
\hline
\end{tabularx}
\end{table*}

\begin{table*}[ht]
\caption{A comparison against leading algorithms in graph classification accuracy}
\label{comp_table}
\centering
\begin{tabularx}{\linewidth}{|l|Y|Y|Y|Y|Y|Y|Y|}
\hline
\textbf{Algorithm}\textbackslash{}\textbf{Dataset} & \multicolumn{1}{c|}{\textbf{MUTAG}} & \multicolumn{1}{c|}{\textbf{PTC}} & \multicolumn{1}{c|}{\textbf{PROTEINS}} & \multicolumn{1}{c|}{\textbf{NCI1}} & \multicolumn{1}{c|}{\textbf{NCI109}} & \multicolumn{1}{c|}{\textbf{D \& D}} & \multicolumn{1}{c|}{\textbf{ENZYMES}} \\ \hline
DCNN{[}2016{]} & 66.98 & 56.60 $\pm$ 2.89 & 61.29 $\pm$ 1.60 & 56.61 $\pm$ 1.04 & 57.47 $\pm$ 1.22 & 58.09 $\pm$ 0.53 & 42.44 $\pm$ 1.76 \\ \hline
PSCN{[}2016{]} & \textbf{88.9 $\pm$ 4.37} & 62.29 $\pm$ 5.68 & 75.00 $\pm$ 2.51 & 76.34 $\pm$ 1.68 & — & — & — \\ \hline
DGCNN{[}2018{]} & 85.83$\pm$1.66 & 58.59 $\pm$ 2.47 & 75.54 $\pm$ 0.94 & 74.44 $\pm$ 0.47 & 75.03 $\pm$ 1.72 & \textbf{79.37 $\pm$ 0.94} & 51.00 $\pm$ 7.29 \\ \hline
GCAPS-CNN{[}2018{]} &  & \textbf{66.01 $\pm$ 5.91} & \textbf{76.40 $\pm$ 4.17} & \textbf{82.72 $\pm$ 2.38} & 81.12 $\pm$ 1.28 & 77.62 $\pm$ 4.99 & \textbf{61.83 $\pm$ 5.39} \\ \hline
RW{[}2003{]} & 83.68 $\pm$ 1.66 & 57.85 $\pm$ 1.30 & 74.22 $\pm$ 0.42 & \textgreater 1 Day & \textgreater 1 Day & \textgreater 1 Day & 24.16 $\pm$ 1.64 \\ \hline
SP{[}2005{]} & 85.79 $\pm$ 2.51 & 58.24 $\pm$ 2.44 & 75.07 $\pm$ 0.54 & 73.00 $\pm$ 0.24 & 73.00 $\pm$ 0.21 & \textgreater 1Day & 40.10 $\pm$ 1.50 \\ \hline
GK{[}2009{]} & 81.58 $\pm$ 2.11 & 57.26 $\pm$ 1.41 & 71.67 $\pm$ 0.55 & 62.28 $\pm$ 0.29 & 62.60 $\pm$ 0.19 & 78.45 $\pm$ 1.11 & 26.61 $\pm$ 0.99 \\ \hline
WL{[}2011{]} & 80.72 $\pm$ 3.00 & 57.97 $\pm$ 0.49 & 74.68 $\pm$ 0.49 & \textbf{82.19 $\pm$ 0.18} & \textbf{82.46 $\pm$ 0.24} & \textbf{79.78 $\pm$ 0.36} & 52.22 $\pm$ 1.26 \\ \hline
DGK{[}2015{]} & 82.66 $\pm$ 1.45 & 60.08 $\pm$ 2.55 & 75.68 $\pm$ 0.54 & 80.31 $\pm$ 0.46 & 80.32 $\pm$ 0.33 & 73.50 $\pm$ 1.01 & 53.43 $\pm$ 0.91 \\ \hline
MLG{[}2016{]} & 84.21 $\pm$ 2.61 & 63.26 $\pm$ 1.48 & \textbf{76.34 $\pm$ 0.72} & 81.75 $\pm$ 0.24 & 81.31 $\pm$ 0.22 & 78.18 $\pm$ 2.56 & \textbf{61.81 $\pm$ 0.99} \\ \hline
\textit{Nauty + Capsules} & 75.7 $\pm$ 9.47 & 63.9 $\pm$ 6.36 & 72.0 $\pm$ 2.61 & 59.4 $\pm$ 2.16 & 58.0 $\pm$ 2.76 & 77.9 $\pm$ 2.49 & 26.1 $\pm$ 5.15 \\ \hline
\textit{BC + Capsules} & \textbf{88.9 $\pm$ 5.49} & \textbf{69.0 $\pm$ 4.98} & 74.1 $\pm$ 3.24 & 65.9 $\pm$ 1.07 & 58.04 $\pm$ 2.78 & 74.86 $\pm$ 3.27 & 27.0 $\pm$ 8.45 \\ \hline
\end{tabularx}
\end{table*}

\subsection{Experiment 1: Ablation Study and Comparison of Labelling Procedures}

\autoref{cnn_vs_caps} shows the classification accuracy for each dataset with the two versions of the algorithm.
It is evident that the model using the Capsule Network (our contribution) outperforms the Patchy-Sans inspired CNN model \cite{niepert2016learning} on all of the datasets, thus provides evidence to support our hypothesis. We also see that in 6 out of 7 of these datasets, the Betweeness Centrality labelling procedure for the ordering and selection of nodes gives better (with respect to how well the classes are separated by the downstream classification algorithms) graph tensor representations than the canonical labelling.

We also observe the effect of the large difference in the number of graphs ($\mathcal{|G|}$) and number of nodes ($N$) in the graphs on the computation time of the algorithms. The smaller graph dataset in both $|\mathcal{G}|$ and $N$ is MUTAG, and the largest are NCI109 and D$\&$D. The time complexities of both algorithms presented here depend on both $\mathcal{|G|}$ and $N$ so the computation time can be largely different. These differences are presented in table \ref{time_tab}. 
As expected, the Capsule Network takes more time to train given that it has a larger number of parameters.

\subsection{Experiment 2: Comparison against Current State of the Art Methods}

Table \ref{comp_table} displays the results found for the two different labelling procedures against the state of the art methods in terms of classification accuracy.

Despite a very modest hyper-parameter search, our results show leading performance in 2 out of the 7 datasets. The datasets where we demonstrate the most competitive results are the ones that have between $0$ and $30$ different node labels (MUTAG, PTC, PROTEINS). However, when the number of node labels is higher (NCI, D\&D, ENZYMES) the algorithm has a lower performance in terms of classification accuracy.

\subsection{Experiment 3: Assessment of the Representational Power of the Capsule Network}
This section presents experiment results on how the explicit tensor representations, the CNNs, and the Capsule Networks encode graph representations corresponding to to the intrinsic structure and the features of graph data.

\begin{table*}[ht]
\centering
\caption{Assessing the representation power of the models}
\begin{tabularx}{\linewidth}{|X|Y|Y|Y|Y|}
\hline
 & \multicolumn{2}{c|}{\textbf{PROTEINS}} & \multicolumn{2}{c|}{\textbf{MUTAG}} \\ \hline
\textbf{Representation Layer} & \multicolumn{1}{c|}{\textbf{Intra-cluster distance}} & \multicolumn{1}{c|}{\textbf{Inter-cluster distance}} & \multicolumn{1}{c|}{\textbf{Intra-cluster distance}} & \multicolumn{1}{c|}{\textbf{Inter-cluster distance}} \\ \hline
Patchy-Sans & 1804.56 & 3.39 & 817.21 & 675.54 \\ \hline
CNN & 9.77 & 45.93 & 133.70 & 1400.57 \\ \hline
Capsule & 10.72 & 1.71 & 126.12 & 514.47 \\ \hline
\end{tabularx}
\label{rep_table}
\end{table*}

\begin{figure}[t]
    \centering
    \begin{subfigure}[c]{0.5\linewidth}
        \includegraphics[width=\linewidth]{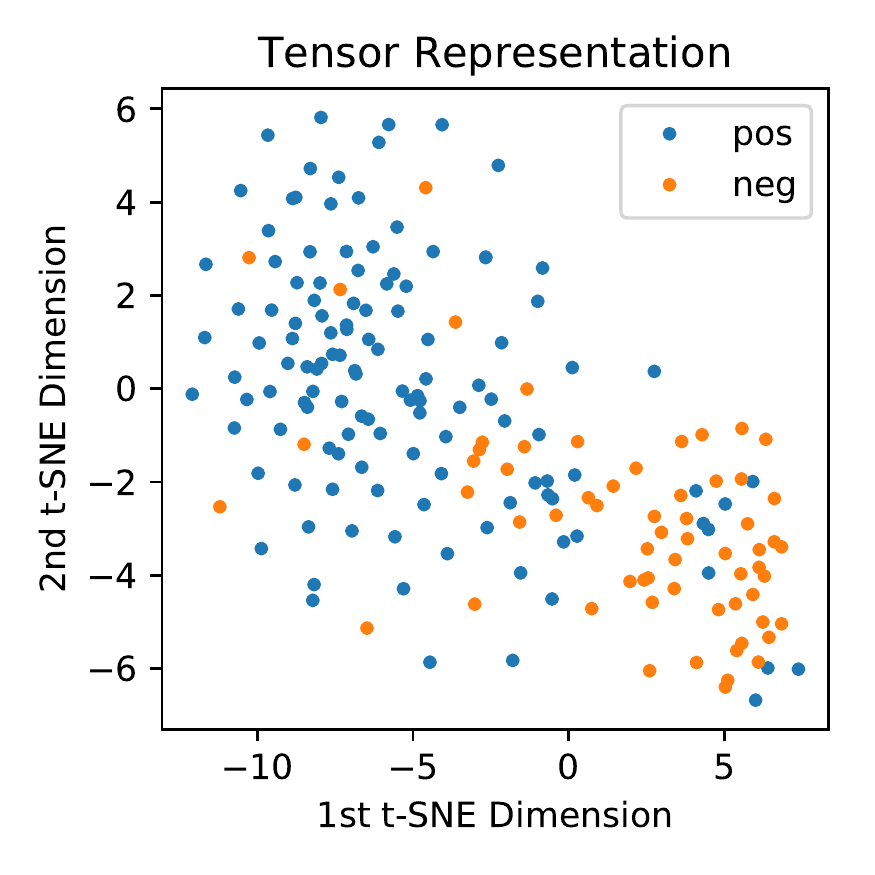}
    \end{subfigure}%
    \begin{subfigure}[c]{0.5\linewidth}
        \includegraphics[width=\linewidth]{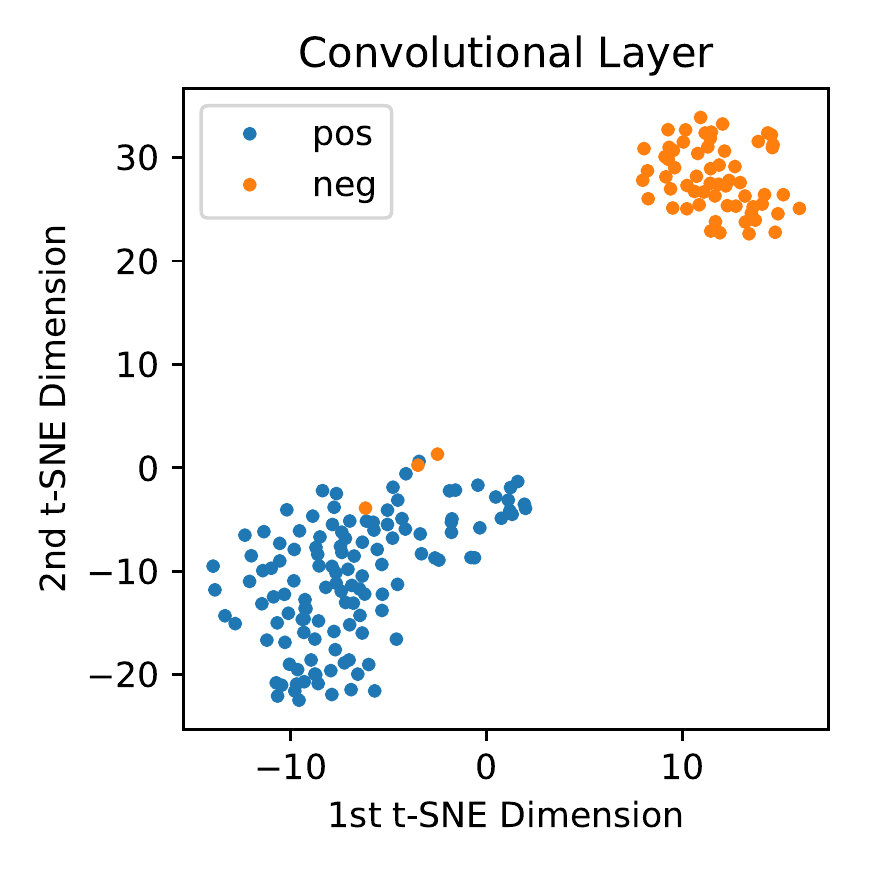}
    \end{subfigure}
    \begin{subfigure}[c]{\linewidth}
        \includegraphics[width=\linewidth]{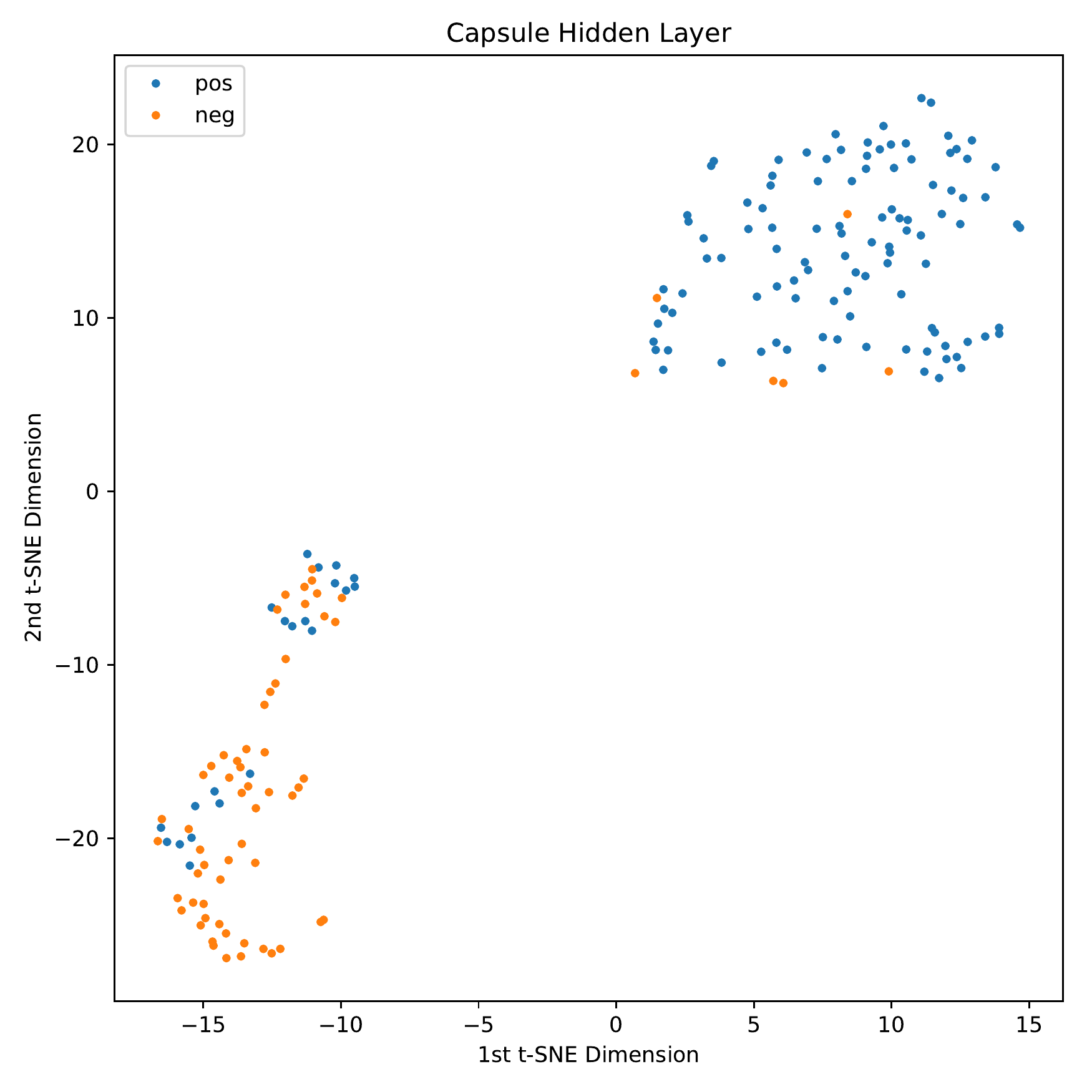}
    \end{subfigure}
    \caption{Mutag t-SNE representations. Top right: CNN Inner Layer. Top left: Tensor Representation. Bottom: Capsule Network inner layer}
    \label{tsne_mutag}
\end{figure}

\begin{figure}[t]
    \centering
    \begin{subfigure}[c]{0.5\linewidth}
        \includegraphics[width=\linewidth]{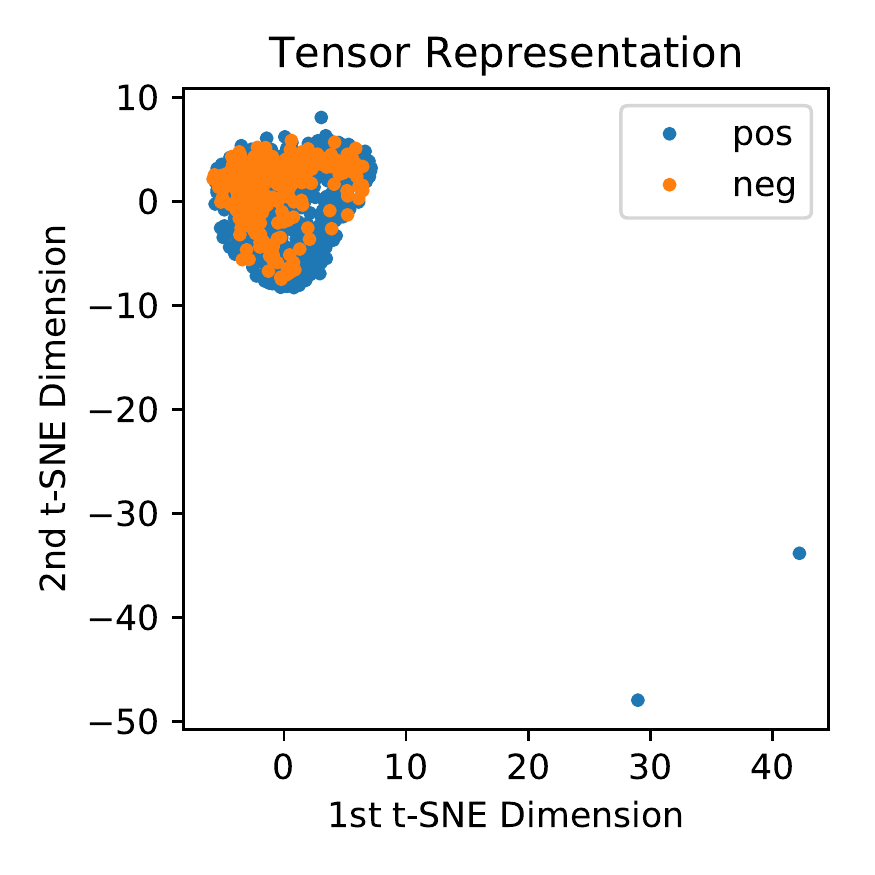}
    \end{subfigure}%
    \begin{subfigure}[c]{0.5\linewidth}
        \includegraphics[width=\linewidth]{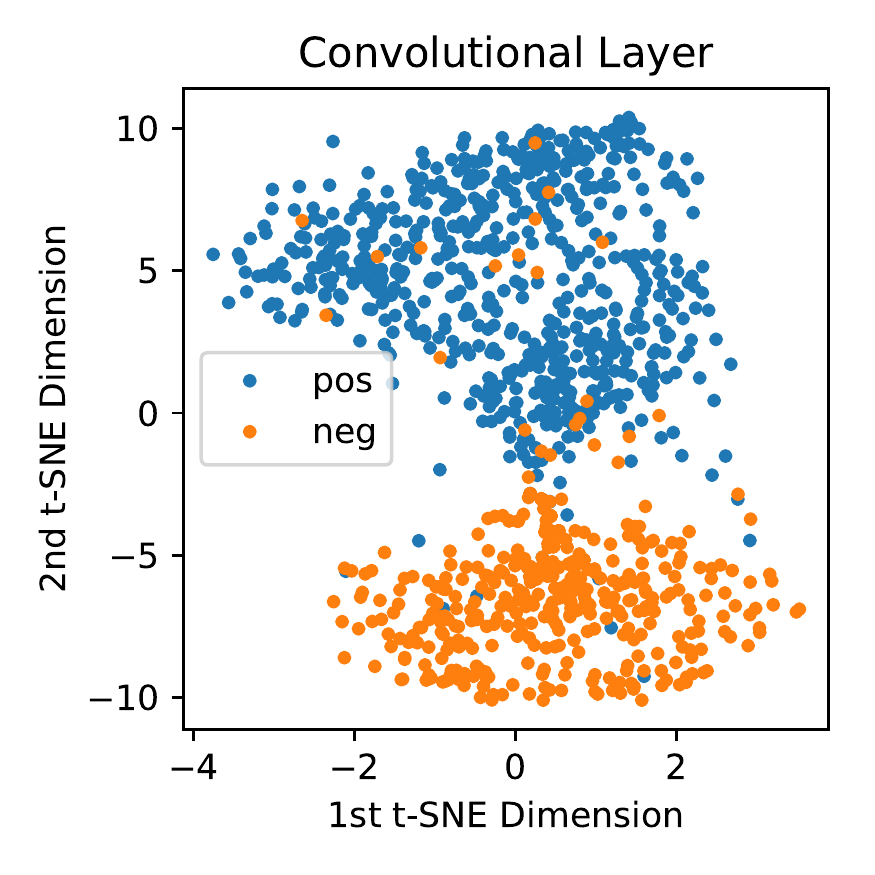}
    \end{subfigure}
    \begin{subfigure}[c]{\linewidth}
        \includegraphics[width=\linewidth]{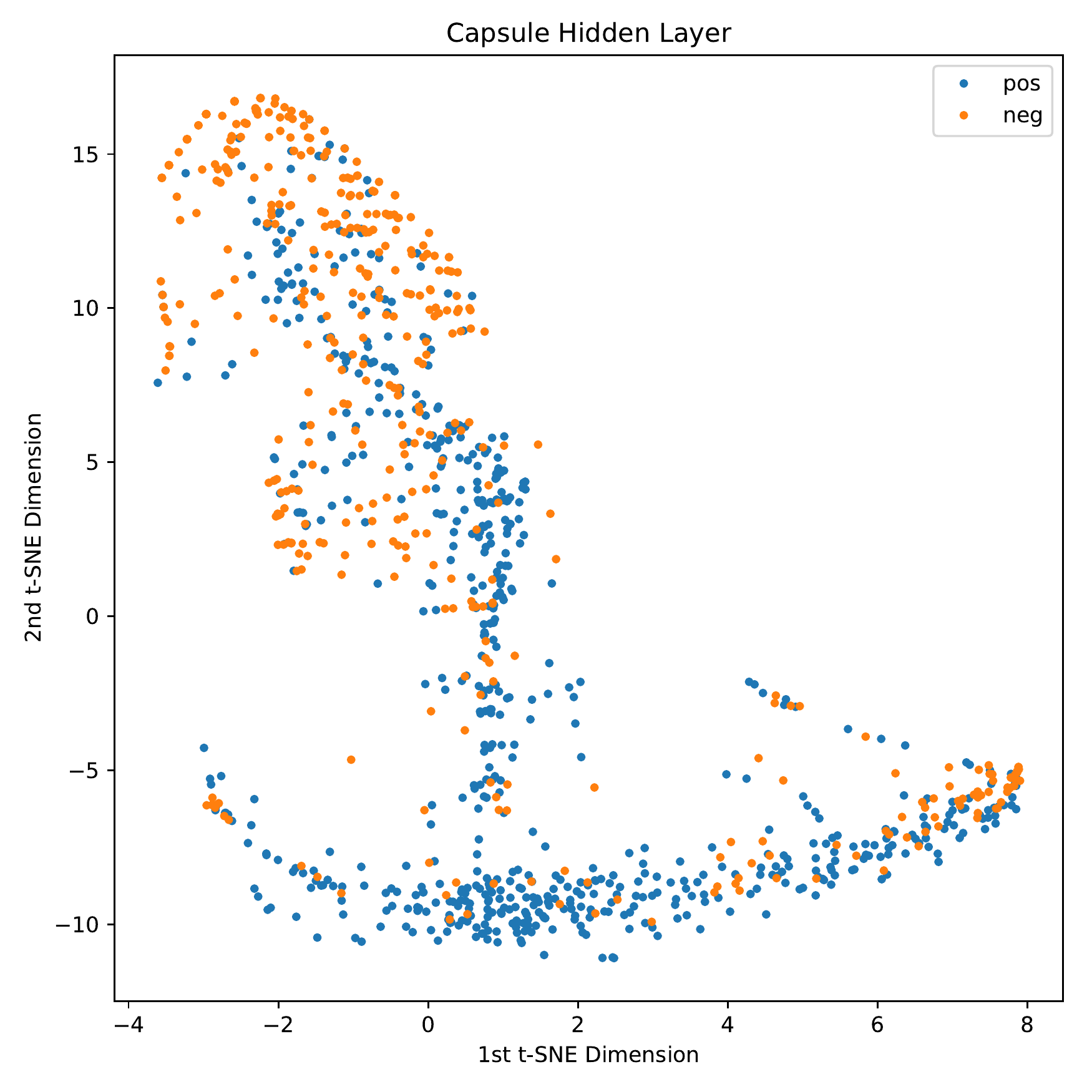}
    \end{subfigure}
    \caption{Proteins t-SNE representations. Top right: CNN inner Layer. Top left: Tensor Representation. Bottom: Capsule Network inner layer}
    \label{tsne_proteins}
\end{figure}

Using trained networks to process the \textit{Proteins} and \textit{Mutag} datasets, we collected the vectors corresponding to the intermediate layer of neurons / capsules (the capsule layer before the first routing operation and the counterpart layer in the standard CNN).
We then apply the manifold embedding algorithm t-SNE \cite{maaten2008visualizing} to render the learned representations
into $\mathcal{R}^2$. \autoref{tsne_proteins} illustrates the t-SNE $\mathcal{R}^2$ embeddings given by the Patchy Sans algorithm (top left), the CNN (top right) and the Capsule Network (bottom). 
The only parameter that needed to be determined is the perplexity, which can be interpreted as a smooth measure of the effective number of neighbors used for the optimization. 
For this experiment we used a perplexity of 10 for the Mutag dataset and of 200 for the Proteins dataset. This values were chosen following the discussion in \cite{maaten2008visualizing}, where it is stated that the performance of t-sne is robust to changes in the perplexity, with typical values between 5 and 50.

In \autoref{tsne_mutag} and \autoref{tsne_proteins}, it is possible to see that the CNN representation appears to better separate the classes than the Capsule Network one, however the classification accuracy of the Capsule Network is significantly higher. One possible reason for this behaviour is that the primary caps layer used for assessing the Capsule Network has not passed through the routing process. This procedure, that operates between this layer and the graph caps layer, would be able to more accurately classify the graphs even if the intermediate representation does not look as clearly separable as the CNN inner layer one.

We can see in \autoref{rep_table} that the CNN produces a better representation than the capsule and the tensor extraction phase alone in terms of separating the positive and negative examples into separate clusters. We quantify this observation with the intra-cluster and inter-cluster distances. The intra-cluster measure is defined as the mean square distance from each point belonging to one class to the center of that class. The inter-class distance is defined as the distance between the center points of each class.

\section{Conclusion and Future work}
We have tested the hypothesis that Capsule Networks are better suited to the permutation invariance problem of graph classification than CNNs when operating on explicit graph tensor representations produced by two labelling procedures. In doing so we have presented and analysed a model for tackling this problem for sets of undirected graphs with discrete node labels of varying numbers of classes.

Our results demonstrate that the Capsule Network indeed outperforms the CNN classifier at this task on all 7 of the benchmark datasets, while also indicating that the use of Betweenness Centrality to inform node ordering and selection for the generation of explicit graph tensor representations is superior to the NAUTY canonical labelling procedure \cite{mckay2014practical} in 6 out of 7 of the datasets.

Although the Capsule Network performs better than the CNN, due to the vastly greater number of parameters to be learned, it also requires a larger execution time. In our experiments we found that on average, the Capsule Network is approximately 8 times slower than the CNN.

We have shown that the Capsule Network with the Betweeness Centrality labelling procedure for node ordering and selection achieves state-of-the-art classification performance on the MUTAG and the PTC datasets. However, on the rest of the datasets, which have a larger number of categorical node features, it is less competitive with these current state-of-the-art methods. We note here, however, that we performed a very limited hyper-parameter search, and do not rule out the possibility that with further search, our model's performance could be significantly improved.

For future work we wish to investigate in detail why our model behaves less well with these datasets. It would also be interesting to try different labelling procedures or perhaps a combination of procedures to investigate the potential further improvement on the model's performance, and of course improving the computational costs of training Capsule Networks is an open area for further work.

\section*{Acknowledgment}


We wish to thank Kyohei Koyama for his assistance in the implementation of our CNN baseline and data preparation with the benchmark datasets. 


\bibliographystyle{IEEEtran}

\end{document}